\documentclass[final]{cvpr}

\usepackage{times}
\usepackage{epsfig}
\usepackage{graphicx}
\usepackage{amsmath}
\usepackage{amssymb}

\usepackage{subcaption}
\usepackage{color}
\usepackage{multirow}
\usepackage[title]{appendix}

\makeatletter
\def\thickhline{%
  \noalign{\ifnum0=`}\fi\hrule \@height \thickarrayrulewidth \futurelet
   \reserved@a\@xthickhline}
\def\@xthickhline{\ifx\reserved@a\thickhline
               \vskip\doublerulesep
               \vskip-\thickarrayrulewidth
             \fi
      \ifnum0=`{\fi}}
\makeatother

\newlength{\thickarrayrulewidth}
\usepackage[pagebackref=true,breaklinks=true,colorlinks,bookmarks=false]{hyperref}

\def\cvprPaperID{2222} 
\def\confYear{CVPR 2021}

\begin{document}


\title{Equalization Loss v2: \\ A New Gradient Balance Approach for Long-tailed Object Detection}

\author{Jingru Tan$^1$ \quad Xin Lu$^2$ \quad Gang Zhang$^3$ \quad  \\ \quad Changqing Yin$^1$ \quad Quanquan Li$^2$\\ 
$^1$Tongji University \quad $^2$SenseTime Research \quad $^3$Tsinghua University  \\
{\tt\small \{tjr120,yinchangqing\}@tongji.edu.cn, \{luxin,liquanquan\}@sensetime.com} \\ 
{\tt\small zhang-g19@mails.tsinghua.edu.cn}
}

\maketitle

\begin{abstract}


Recently proposed decoupled training methods emerge as a dominant paradigm for long-tailed object detection. But they require an extra fine-tuning stage, and the disjointed optimization of representation and classifier might lead to suboptimal results. However, end-to-end training methods, like equalization loss (EQL), still perform worse than decoupled training methods. In this paper, we reveal the main issue in long-tailed object detection is the imbalanced gradients between positives and negatives, and find that EQL does not solve it well. To address the problem of imbalanced gradients, we introduce a new version of equalization loss, called equalization loss v2 (EQL v2), a novel gradient guided reweighing mechanism that re-balances the training process for each category independently and equally. Extensive experiments are performed on the challenging LVIS benchmark. EQL v2 outperforms origin EQL by about 4 points overall AP with 14 $\sim$ 18 points improvements on the rare categories. More importantly, it also surpasses decoupled training methods. Without further tuning for the Open Images dataset, EQL v2 improves EQL by 7.3 points AP, showing strong generalization ability. Codes have been released at \url{https://github.com/tztztztztz/eqlv2}


\end{abstract}

\begin{figure}[t]
\begin{center}
\includegraphics[width=\linewidth]{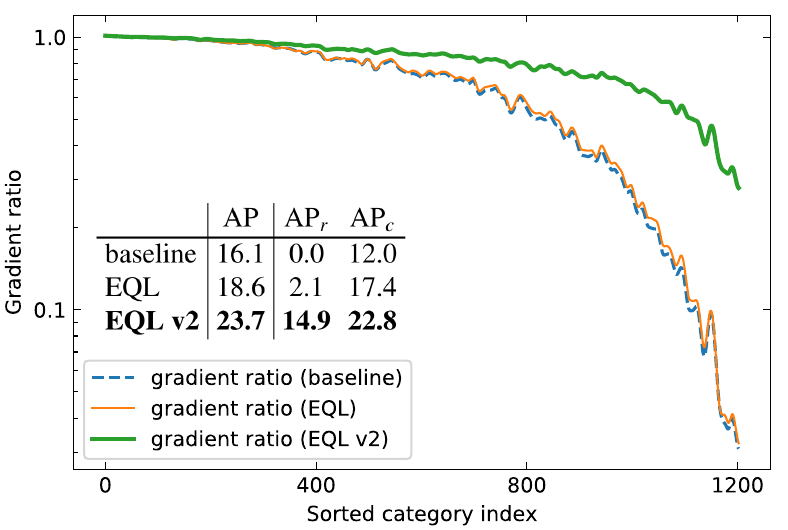}
\end{center}
  \caption{Visualization of accumulative gradients ratio of different trained models. Best view in color. The x-axis is the sorted category index of 1203 categories of LVIS dataset. The y-axis is the accumulative gradient ratio of positives to negatives. Here gradient is the gradient of the output logits with respect to classification loss. AP\textsubscript{\textit{r}} and AP\textsubscript{\textit{c}} are the AP for rare and common categories. }
\label{fig:comparison_eql}
\end{figure}

\section{Introduction}

Object detection is a fundamental computer vision task that aims to recognize and locate objects of a set of pre-defined categories. 
Modern object detectors~\cite{ren2015faster, redmon2016you, liu2016ssd, lin2017feature, lin2017focal, cai2018cascade} have shown promising results on some conventional benchmarks such as COCO~\cite{lin2014microsoft} and PASCAL VOC~\cite{everingham2010pascal}. Collected images in these datasets have been carefully selected and the quantities of each category are relatively balanced. However, in natural images, quantities of categories subject to a long-tailed Zipfian distribution.
It means that, in a realistic scenario, we are confronted with a more complex situation that the obtained objects show an extreme imbalance in different categories.

The difficulty of training detectors on a long-tailed dataset mainly comes from two aspects. First, deep learning methods are hungry for data, but annotations of tail classes (classes with few samples) might be insufficient for training. Second, the model tends to bias towards head classes (classes with many samples) since the head class objects are the overwhelming majority in the entire datasets.

Current state-of-the-art approaches are based on decoupled training schema~\cite{kang2019decoupling, li2020overcoming, wang2020devil}. In general, decoupled training involves a two-stage pipeline that learns representations under the imbalance dataset at the first stage, then re-balances the classifier with frozen representation at the second stage. Despite the success of the decoupled training, it needs an extra fine-tuning stage in training phase. In addition, the representation could be suboptimal since it is not jointly learned with the classifier. So a natural question to ask is: could end-to-end training methods match or surpass the accuracy of decoupled training methods?

Recently, Tan \emph{et~al.}~\cite{tan2020equalization} propose the Equalization Loss (EQL)~\cite{tan2020equalization}, an end-to-end re-weighing loss function, to protect the learning of tail categories by blocking some negative gradients. Although EQL makes improvements to long-tailed training, the accuracy gap between the end-to-end and decoupled training approaches still exists. To take a step forward, we analyze the gradient statistics of EQL. Here we plot the positive gradient to negative gradient ratio accumulated in the entire training process for each category classifier, as present in Figure~\ref{fig:comparison_eql}. The key observation is: for head categories, the ratio is close to 1, which means the positive gradients and the negative gradients have similar magnitude; for tail categories, the gradients are near 0, which means the positive gradients are overwhelmed by the negative gradients. Therefore the gradient ratio could indicate whether a classifier is trained in balance. Compared with the baseline (blue line), the gradient ratio of EQL (orange line) just increases slightly. 

In this paper, we propose a new version of equalization loss, called equalization loss v2 (EQL v2) which improves the long-tailed object detection by balancing the positive to negative gradient ratio. In EQL v2, we first model the detection problem as a set of independent sub-tasks, each task for one category. Next, we propose a gradient guided re-weighing mechanism to balance the training process of each task independently and equally. Specifically, the accumulated gradient ratio is used as an indicator to up-weight the positive gradients and down-weight the negative gradients. It dynamically controls the training of all sub-tasks and each sub-task is treated equally with the same simple re-weighing rule. The positive to negative gradient ratio of EQL v2 are shown in Figure~\ref{fig:comparison_eql} (green line). Compared to the baseline and EQL, EQL v2 achieves a more balanced training for most categories. 

We conduct experiments on two long-tailed object detection dataset, LVIS~\cite{gupta2019lvis} and OpenImages~\cite{kuznetsova2018open}. On LVIS, compared to the baseline models, including Mask R-CNN~\cite{he2017mask} and Cascade Mask R-CNN~\cite{cai2018cascade}, it increases overall AP by about 6 points and gains 17 $\sim$ 20 points AP for tail categories. It outperforms EQL by about 4 points AP. In addition, EQL v2 surpasses all of the existing long-tailed object detection methods, including end-to-end training and decoupled training methods. On OpenImages, EQL v2 achieves a 9 points AP gain over the baseline model with the same hyper-parameters as on LVIS, which shows the good generalization ability.

\section{Related Work}

\noindent \textbf{General object detection.} Modern object detection frameworks rely on the outstanding ability of classification powered by convolutional neural networks (CNN) \cite{simonyan2014very, szegedy2015going, he2016deep}. They can be divided into region-based detectors \cite{girshick2014rich, girshick2015fast, ren2015faster, lin2017feature, lin2017focal, he2017mask, cai2018cascade, chen2019hybrid} and anchor-free detectors~\cite{tian2019fcos, zhu2019feature, kong2020foveabox, law2018cornernet, duan2019centernet, zhou2019bottom} depending on the concept they want to classify. However, all those frameworks are developed under the condition of balanced data. When it comes to the long-tailed distribution of data, the performance deteriorates severely due to the imbalanced among categories.


\vspace{3mm} \noindent \textbf{Long-tailed image classification.} Common solutions for long-tailed image classification are data re-sampling and loss reweighing. However, data re-sampling \cite{shen2016relay, mahajan2018exploring, han2005borderline, chawla2002smote} methods have to access the pre-computed statistics of data distribution and might make models under the risks of over-fitting for tail classes and under-fitting for head classes. For the loss re-weighing methods, including instance-level~\cite{lin2017focal, li2019gradient, shu2019meta} ones and class-level ones~\cite{cui2019class, wang2017learning, huang2016learning, zhang2017range, cao2019learning}, they suffer from the sensitive hyper-parameters, the optimal setting for different dataset might vary largely and finding them takes too many efforts. There are also some works trying to transfer the knowledge from head classes to tail classes. OLTR~\cite{liu2019large} designs a memory module to augment the feature for tail classes. \cite{chu2020feature, yin2019feature} augment the under-represented classes in the feature space by using the knowledge learned from head classes. Recently, the decoupled training \cite{kang2019decoupling} schema attracts much attention. They argue that universal representations can be learned without re-sampling, and the classifier should be re-balanced in the second fine-tuning stage with representations frozen. In spite of excellent results, the extra fine-tuning stage seems unnatural and we can not explain why the representation and the classifier have to be learned separately.


\vspace{3mm} \noindent \textbf{Long-tailed object detection.} Long-tailed object detection is a more difficult problem than long-tailed classification. It has to find all objects with various scale in every possible location. Li \emph{et~al.}~\cite{li2020overcoming} empirically find methods that are designed for long-tailed image classification can not achieve good results in object detection. Tan \emph{et~al.}~\cite{tan2020equalization} first shows the tail classes are heavily suppressed by the head classes and they propose an equalization loss to tackle this problem by ignoring the suppressing part for tail categories. However, they think the negative suppressing comes from competition of foreground categories and ignore the impact of background proposal. Instead, we treat background and foreground uniformly. EQL also has to access the frequency of categories and uses a threshold function to explicitly split head and tail categories. LST~\cite{hu2020learning} models the learning for long-tailed distribution as a kind of incremental learning, the learning switches from head classes to tail classes in several cascaded stages. SimCal~\cite{wang2020devil} and Balanced GroupSoftmax (BAGS) \cite{li2020overcoming} follow the spirit of decoupled training. For SimCal, they train an extra classification branch with class-balanced proposals in the fine-tuning stage and combine its score with a normal trained softmax classification branch via dual inference. BAGS divides all categories into several groups based on the instance count and does softmax separately in each group to avoid the domination of head classes. In contrast, our method does not have to split categories into different groups and treat all categories equally. Moreover, we do not need the fine-tuning stage and can be trained end-to-end. Tang \emph{et~al.}~\cite{tang2020long} shows that SGD momentum makes the classifier biased towards head classes. They introduce causal intervention in training and remove the biased part for tail classes in inference. On the contrary, our method is simpler and more efficient, and keeps consistent between training and inference.

\section{Equalization Loss v2}

In this section, we introduce the Equalization Loss v2. We begin by revisiting the entanglement of instances and tasks in Section~\ref{sec:entangle}, then present our novel gradient guided reweighing strategy in Section~\ref{sec:ggr}. 


\subsection{Entanglement of Instances and Tasks}
\label{sec:entangle}

Suppose we have a batch of instances $\mathcal{I}$ and their representations. To output logits $\mathcal{Z}$ for $C$ categories, a weight matrix $\mathcal{W}$ is used as a linear transformation of representations. Each weight vector in $\mathcal{W}$, which we refers as a category classifier, is responsible for a specific category, \ie a task. Then the output logits are transformed to an estimated probability distribution $\mathcal{P}$ by the sigmoid function. We expect that for each instance, only the corresponding classifier gives the high score while others give a low score. That is saying, \textit{one task with positive label and $C - 1$ tasks with negative labels are introduced by a single instance}. Hence, we can calculate the actual number of positive samples $m_{j}^{pos}$ and negative samples $m_{j}^{neg}$ for classifier $j$:

\begin{equation}
    m_{j}^{pos} = \sum_{i \in \mathcal{I}} y_{j}^{i}, \quad m_{j}^{neg} = \sum_{i \in \mathcal{I}} (1 - y_{j}^{i})
\label{eq:pos_neg_sample_number}
\end{equation}

Where the $y^{i}$ is the one-hot groud truth label for the $i$-th instance, and usually we have $\sum_{j}y_{j}^{i} = 1$. The ratio of expectation of positive samples to the negative samples over the dataset is then:

\begin{equation}
    \frac{\mathbb{E}|m_{j}^{pos}|}{\mathbb{E}|m_{j}^{neg}|} = \frac{1}{\frac{N}{n_j} - 1} 
\label{eq:pos_neg_number_ratio}
\end{equation}

Where $n_j$ is the instance number of category $j$ and $N$ is the total instance number over the dataset. Equation \ref{eq:pos_neg_number_ratio} shows that if we consider each classifier separately, the ratio of the positive samples to the negative samples could have a big difference for different classifiers. 


\subsection{Gradient Guided Reweighing}
\label{sec:ggr}

Obviously, we have $ \mathbb{E}|m_{j}^{pos}| \ll \mathbb{E}|m_{j}^{neg}|$ especially when category $j$ is a rare category. But the ratio in Equation~\ref{eq:pos_neg_number_ratio} might not be a good indicator of how balanced the training is. The reason behind it is that the influence of each sample is different. For example, the negative gradients accumulated by large quantities of easy negatives might be smaller than the positive gradients generated by a few hard positives. Therefore, we directly choose gradient statistics as a metric to indicate whether a task is in balanced training. The positive and negative gradients for each classifier's output $z_{j}$ with respect to the loss $\mathcal{L}$ are formulated as:

\begin{equation}
    \nabla_{z_{j}}^{pos} (\mathcal{L}) = \frac{1}{|\mathcal{I}|} \sum_{i \in \mathcal{I}} y_{j}^{i}(p_{j}^{i} - 1)
\end{equation}

\begin{equation}
    \nabla_{z_{j}}^{neg} (\mathcal{L}) = \frac{1}{|\mathcal{I}|} \sum_{i \in \mathcal{I}} (1 - y_{j}^{i})p_{j}^{i}
\end{equation}

 $p_{j}^{i}$ is the estimated probability of category $j$ for the $i$-th instance. The basic idea of gradient guided balanced reweighing is that we up-weight the positive gradients and down-weight negative gradients for each classifier independently according to their \textit{accumulated} gradient ratio of positives to negatives.
 
To achieve this, we first define $g_{j}^{(t)}$ as the ratio of \textit{accumulated} positive gradients to negative gradients of task $j$ until the iteration $t$. Then the weight for positive gradients $q_{j}^{t}$ and negative gradients $r_{j}^{t}$ at this iteration can be computed by:

 \begin{equation}
     q_{j}^{(t)} = 1 + \alpha (1 - f(g_{j}^{(t)})), \quad  r_{j}^{(t)} = f(g_{j}^{t}) 
 \end{equation}
 
 \noindent Where $f(\mathord{\cdot})$ is a mapping function:
 
 \begin{equation}
     f(x) = \frac{1}{1 + e^{-\gamma(x - \mu)}}
 \label{eq:map_func}
 \end{equation}

\noindent After obtaining $q_{j}^{t}$ and  $r_{j}^{t}$, we apply them to the positive gradient and negative gradient for the current batch, so the re-weighted gradients become:

 \begin{equation}
    \nabla_{z_{j}}^{pos'} (\mathcal{L}^{(t)}) =  q_{j}^{(t)}\nabla_{z_{j}}^{pos} (\mathcal{L}^{(t)})
\end{equation}

 \begin{equation}
 \nabla_{z_{j}}^{neg'} (\mathcal{L}^{(t)}) = r_{j}^{(t)} \nabla_{z_{j}}^{neg} (\mathcal{L}^{(t)})
\end{equation}

\noindent Finally we update the ratio of \textit{accumulated} positive gradients to negative gradients for the next iteration $t+1$:

 \begin{equation}
     g_{j}^{t+1} = \frac{\sum_{t^{*}=0}^{t}|\nabla_{z_{j}}^{pos'} (\mathcal{L}^{(t^{*})})|}{\sum_{t^{*}=0}^{t}|\nabla_{z_{j}}^{neg'} (\mathcal{L}^{(t^{*})})|}
 \end{equation}

\begin{table*}
   \centering
   \setlength\tabcolsep{12pt}
   \begin{tabular}{l | c c | c c c c c}
      method & \#sampler & \#epoch & AP & AP\textsubscript{\textit{r}} &
      AP\textsubscript{\textit{c}} & AP\textsubscript{\textit{f}} & AP\textsubscript{\textit{b}}\\
      \thickhline
      \textit{End-to-end Training} &  &  &  & &  &  \\
      (a)~Softmax & random & 12 & 16.1 & 0.0 & 12.0 & 27.4 & 16.7 \\
      (b)~Sigmoid & random & 12 & 16.5 & 0.0 & 13.1 & 27.3 & 17.2 \\
      (c)~EQL~\cite{tan2020equalization} & random & 12 & 18.6 & 2.1 & 17.4 & 27.2 & 19.3 \\
      (d)~RFS~\cite{gupta2019lvis} & repeat factor & 12 & 22.2 & 11.5 & 21.2 & 28.0 & 22.9 \\
      \hline
      \textit{Decoupled Training} &  &  &  & &  &  \\
      (e) LWS~\cite{kang2019decoupling} & random/balance & 12+12 & 17.0 & 2.0  & 13.5 & 27.4 & 17.5 \\
      (f) cRT~\cite{kang2019decoupling}& random/balance & 12+12 & 22.1 & 11.9 & 20.2 & 29.0 & 22.2 \\
      (g) BAGS~\cite{li2020overcoming} & random/random & 12+12 & 23.1 & 13.1 & 22.5 & 28.2 & 23.7 \\
      \hline
      (h) EQL v2 (Ours) & random & 12 & 23.7 & 14.9 & 22.8 & 28.6 & 24.2 \\
   \end{tabular}
      \caption{
      Comparison with end-to-end and decoupled training methods on LVIS v1.0 \texttt{val} set with ResNet-50-FPN Mask R-CNN by 1x schedule. For cRT and LWS, they use class-balance sampler to fine-tune their model in the second stage, and BAGS uses a random sampler following the origin paper. Instead, our method train models in an end-to-end fashion without any fine-tuning stage. }
   \label{tab:compare_e2e_decouple}
\end{table*}

\begin{table}
   \centering
   \setlength\tabcolsep{5pt}
   \begin{tabular}{c c c | c c c c c}
         obj? & neg? & pos? & AP & AP\textsubscript{\textit{r}} & AP\textsubscript{\textit{c}} & AP\textsubscript{\textit{f}} & AP\textsubscript{\textit{b}} \\
         \thickhline
         & & & 16.1 & 0 & 12.0 & 27.4 & 17.2 \\
         \checkmark & & & 18.1 & 1.9 & 16.4 & 28.3 & 19.0\\
         \checkmark & \checkmark & & 19.7 & 7.3 & 17.6 & 27.6 & 20.5 \\
         \checkmark & \checkmark & \checkmark & \textbf{23.7} & \textbf{14.9} &\textbf{22.8} & 28.6 & \textbf{24.2} \\
         
   \end{tabular}
   \caption{Effect of different components. \textit{obj} for adding the category-agnostic task, \textit{neg} for reweighing negative gradients, \textit{pos} for reweighing positive gradients. Models are trained with random samplers by standard 1x schedule. AP and AP\textsubscript{\textit{b}} denotes mask AP and box AP respectively.}
   \label{tab:component_analysis}
\end{table}

\section{Experiments}

\subsection{Dataset and Evaluation Metric}

LVIS~\cite{gupta2019lvis} is a new benchmark for long-tailed object recognition. It provides precise bounding box and mask annotations for various categories with long-tailed distribution. We mainly perform experiments on the recently released challenging LVIS v1.0 dataset. It consists of 1203 categories. We train our models on the \texttt{train} set, which contains about 100k images with 1.3M instances. In addition to widely-used metric AP across IoU threshold from 0.5 to 0.95, LVIS also reports AP\textsubscript{r} (rare categories with 1-10 images), AP\textsubscript{c} (common categories with 11-100 images), AP\textsubscript{f} (frequent categories with $>$ 100 images). Since LVIS is federated dataset, categories are not annotated exhaustively. Each image have two more types of labels: \texttt{pos\_category\_ids} and \texttt{neg\_category\_ids}, indicating which categories are or are not present in that image. Detection results that do not belong to those categories will be ignored for that image. We report results on the \texttt{val} set of 20k images.

\subsection{Implementation Details}

We implement our method using MMDetection~\cite{chen2019mmdetection}. Models are trained using SGD with a momentum of 0.9 and a weight decay of 0.0001. The ResNet~\cite{he2016deep} backbones are initialized by ImageNet pre-trained models. Following the convention, scale jitter and horizontal flipping are used in training and no test time augmentation is used. We use a total batch size of 16 on 16 GPUs (1 image per GPU), and set the initial learning rate to 0.02. Following \cite{gupta2019lvis}, the maximum number of detection per image are up to 300, and the minimum score threshold are reduced to 0.0001 from 0.01. For our method, we set $\gamma = 12$, $\mu = 0.8$ and $\alpha = 4$. More details about hyper-parameters are showcased in section~\ref{sec:hyper_param}. Since EQL v2 uses sigmoid loss function, we initialize the bias of the last fully-connected classification layer(fc-cls) with values of 0.001 to stabilize the training at the beginning. 

Following ~\cite{li2020overcoming}, we also add a branch for detecting objectiveness instead of concrete category to reduce false-positives, which we refer as \textbf{category-agnostic task}. In the training phase, this task treats all other tasks' positive samples as its positive samples. In the inference phase, the estimated probability of other sub-tasks becomes:

\begin{equation}
    p'_{j} = p_{j} * p_{obj} 
\end{equation}

Where $p_{obj}$ is the probability for a proposal being a object. The proposed gradient guided reweighing are not applied on the category-agnostic task. 


\subsection{Ablation Studies}

We take Mask R-CNN~\cite{he2017mask} equipped with ResNet-50 and FPN~\cite{lin2017feature} as our baseline model. The effect of each component is shown in Table \ref{tab:component_analysis}. The baseline model performs poorly on the tail classes, resulting in 0\% and 12.0\% AP for rare and common categories. And the performance gap between head and tail classes are very large. Adding a category-agnostic task helps all categories to some extent, improving the overall AP by 2.0\% but not very much for the rare categories since the main problem for them is the unbalanced positive and negative gradients, \ie. their positive gradients are overwhelmed by negative gradients cause by a vast number of negative samples. By down-weighting the influence of negative gradients, their accuracy is boosted significantly (5.4\% for rare categories).  Up-weighting the positive gradient helps to achieve a more balanced ratio of positive to negative gradients. It brings a 7.6\% performance boosting for rare categories. With these three components, we achieve a 23.7\% AP, outperforming the baseline model 16.1\% AP by a large margin without any re-sampling techniques. These ablation experiments verified the effectiveness of our proposed loss function.

\begin{table}
   \centering
   \setlength\tabcolsep{10pt}
   \begin{tabular}{l | c c c c}
         method & AP & AP\textsubscript{\textit{r}} & AP\textsubscript{\textit{c}} & AP\textsubscript{\textit{f}} \\
         \thickhline
         (a)~Mask-R50 & 20.5 & 2.0 & 19.0 & 30.3 \\
         (b)~$+$EQL v2 & 26.2 & 19.1 & 25.0 & 30.7 \\
         \hline
         (c)~Mask-R101 & 21.7 & 1.6 & 20.7 & 31.7 \\
         (d)~$+$EQL v2 & 27.5 & 20.5 & 26.2 & 32.0 \\
   \end{tabular}
   \caption{Results of larger backbones with a longer 3x schedule. A Random sampler is used. The models are trained with totally 36 epochs, and the learning rate is divided by 10 at the 28th epoch and 34th epoch respectively.}
   \label{tab:longer_schedule}
\end{table}

\begin{table*}
   \centering
   \begin{tabular}{l |c  c | c c c c c }
      method & framework & \#sampler & AP & AP\textsubscript{\textit{r}} & AP\textsubscript{\textit{c}} & AP\textsubscript{\textit{f}} & AP\textsubscript{\textit{b}} \\
     \thickhline
     RFS\textsuperscript{\textdagger} \cite{gupta2019lvis} & Mask-R50 & repeat & 24.4 & 14.5 & 24.3 & 28.4 & - \\ 
     EQL\textsuperscript{\textdagger} \cite{tan2020equalization} & Mask-R50 & random & 22.8 & 11.3 & 24.7 & 25.1 & 23.3 \\
     LST\textsuperscript{\textdagger} \cite{hu2020learning} & Mask-R50 & - & 23.0 & - & - & - & - \\
     SimCal\textsuperscript{\textdagger} \cite{wang2020devil} & Mask-R50 & random/balance & 23.4 & 16.4 & 22.5 & 27.2 & -\\
     Forest R-CNN\textsuperscript{\textdagger} \cite{wu2020forest} & Mask-R50 & nms-resample& 25.6 & 18.3 & 26.4 & 27.6 & 25.9 \\
     BAGS\textsuperscript{\textdagger} \cite{li2020overcoming} & Mask-R50 & random/random & 26.3 & 18.0 & 26.9 & 28.7 & 25.8 \\
     De-confound-TDE\textsuperscript{\textdagger} \cite{tang2020long} & Cascade-R101 & random &  28.4 & 22.1 & 29.0 & 30.3 & 31.0 \\
     BAGS\textsuperscript{\textdagger}~\cite{li2020overcoming} & HTC-X101 & random/random & 31.2 & 23.4 & 32.3 & 32.9 & 33.7 \\
     \hline
     EQL v2 (Ours) & Mask-R50 & random & 27.1 & 18.6 & 27.6 & 29.9 & 27.0 \\
     EQL v2 (Ours) & Mask-R101 & random & 28.1 & 20.7 & 28.3 & 30.9 & 28.1 \\
     EQL v2 (Ours) & Cascade-R101 & random & 30.2 & 23.0 & 30.9  & 32.1 & 33.0 \\
     EQL v2 (Ours) & HTC-X101 & random & \textbf{32.0} & \textbf{24.2} & \textbf{32.8} & \textbf{34.1} & \textbf{34.0} \\
    \end{tabular}
    \caption{Comparison with state-of-the-art methods on \textbf{LVIS v0.5} \texttt{val} set. \textdagger~indicates that the reported result is directly copied from referenced paper. 'Mask-R50' indicates Mask R-CNN~\cite{he2017mask} with ResNet50-FPN~\cite{he2016deep, lin2017feature}, 'Cascade' is for Cascade Mask R-CNN~\cite{cai2018cascade}, 'HTC' is for Hybrid Task Cascade~\cite{chen2019hybrid}. Models are trained with the corresponding \textbf{LVIS v0.5} \texttt{train} set.}
    \label{tab:comparison_sota_lvis0_5}
\end{table*}

\begin{table*}
   \centering
   \setlength\tabcolsep{10pt}
   \begin{tabular}{l |c  | c c c c c }
      method & framework & AP & AP\textsubscript{\textit{r}} & AP\textsubscript{\textit{c}} & AP\textsubscript{\textit{f}} & AP\textsubscript{\textit{b}} \\
     \thickhline
     De-confound\textsuperscript{\textdagger}~\cite{tang2020long} & Cascade-R101 & 23.5 & 5.2 & 22.7 & 32.3 & 25.8 \\
     De-confound-TDE\textsuperscript{\textdagger}~\cite{tang2020long} & Cascade-R101 &  27.1 & 16.0 & 26.9 & 32.1 & 30.0 \\
     \hline
     Mask R-CNN & Mask-R50 & 19.2 & 0 & 17.2 & 29.5 & 20.0 \\
     EQL\textsuperscript{*} & Mask-R50 & 21.6 & 3.8 & 21.7 & 29.2 & 22.5 \\
     EQL v2 (Ours) & Mask-R50 & 25.5 & 17.7 & 24.3 & 30.2 & 26.1 \\
     \hline
     Mask R-CNN & Mask-R101 & 20.8 & 1.4 & 19.4 & 30.9 & 21.7 \\
     EQL\textsuperscript{*} & Mask-R101 & 22.9 & 3.7 & 23.6 & 30.7 & 24.2 \\
     EQL v2 (Ours) & Mask-R101 & 27.2 & 20.6 & 25.9 & 31.4 & 27.9 \\
     \hline
     Cascade Mask R-CNN & Cascade-R101 & 22.6 & 2.0 & 22.0 & 32.5 & 25.2 \\
     EQL\textsuperscript{*} & Cascade-R101 & 24.5 & 4.1 & 25.8 & 32.0 & 27.2 \\
     EQL v2 (Ours) & Cascade-R101 & \textbf{28.8} & \textbf{22.3} & \textbf{27.8} & \textbf{32.8} & \textbf{32.3} \\
    \end{tabular}
    \caption{Comparison with state-of-the-art methods on \textbf{LVIS v1.0} \texttt{val} set. \textdagger~indicates that the reported result is directly copied from the referenced papers. * indicates our re-implementation. Models are trained using \textbf{LVIS v1.0} \texttt{train} set. We train our models with a standard 2x schedule.}
    \label{tab:comparison_sota_lvis1}
\end{table*}

\subsection{Main Results}

\vspace{1mm} \noindent \textbf{Comparison with Decoupled Training methods.} We mainly compare our method with three decoupled training methods (cRT~\cite{kang2019decoupling}, LWS~\cite{kang2019decoupling}, and BAGS~\cite{li2020overcoming}). The results are present in Table \ref{tab:compare_e2e_decouple}. The decoupled training models (Table~\ref{tab:compare_e2e_decouple}~(e)~(f)~(g)) are first initialized from naive softmax baseline (Table \ref{tab:compare_e2e_decouple}~(a)), then re-train their classifier layer (fc-cls) for another 12 epoch with other layers frozen, resulting in a total 24 epoch training. Those decoupled training methods all improve the AP, mainly for tail classes. The improvements brought by LWS is limited. We conjecture it is because that LWS only learns a scaling factor to adjust the decision boundary of the classifier but the classifier itself is not good and imbalanced. Our method achieves a overall 23.7\% AP, increasing AP\textsubscript{r} by 14.9\%, AP\textsubscript{c} by 10.8\%. It is worth noting that EQL v2 does not require the extra fine-tuning stage, and the representation and classifier are learned jointly. More importantly, it has already surpassed the decoupled training methods. 

\vspace{1mm} \noindent \textbf{Comparison with End-to-End Training methods.} Table~\ref{tab:compare_e2e_decouple} compares EQL v2 with two popular end-to-end training methods, Repeat Factor Sampling~\cite{gupta2019lvis} (re-sampling) and Equalization Loss~\cite{tan2020equalization} (re-weighting). With a random sampler, our method outperforms naive softmax  and EQL by a large margin, 7.6\% and 5.1\% respectively. Note that RFS repeats images that contains tail categories in each epoch, so it increases the total training time. Instead, our method only uses a random sampler and does not increase the training time, and achieves better results, 23.7\% \vs 22.2\%.

\vspace{1mm} \noindent \textbf{Larger Model \& Longer Training.} To verify the generalization ability across different backbones and training schedule. We conduct experiments with larger models by a 3x schedule.  The results are present in Table \ref{tab:longer_schedule}. Note that training Mask R-CNN with longer schedule does not help rare categories a lot (Table \ref{tab:compare_e2e_decouple}~(a) \vs Table \ref{tab:longer_schedule}~(a)), the AP of rare categories is still bad because rare categories are heavily suppressed by the negative gradients caused by the entanglement of instances and tasks. In contrast, with the proposed EQL v2, the performance of rare categories can be further improved from 14.9\% to 19.1\% (Table \ref{tab:compare_e2e_decouple}~(h) and Table \ref{tab:longer_schedule}~(b)). When using large ResNet-101 backbone, we do not observe the over-fitting of tail classes in such a long schedule, and the gap between Mask R-CNN and EQL v2 holds.

\subsection{Comparison with State-of-the-Art Methods}
In this section, we compare our method with other work that report state-of-the-art results on LVIS v0.5 and LVIS v1.0. The results on LVIS v0.5 is present in Table \ref{tab:comparison_sota_lvis0_5}, including RFS~\cite{gupta2019lvis} for re-sampling, EQL~\cite{tan2020equalization} for re-weighing, LST~\cite{hu2020learning} for incremental learning, SimCal~\cite{wang2020devil} and BAGS~\cite{li2020overcoming} for decoupled training, Forest R-CNN~\cite{wu2020forest} for hierachy classification, De-cofound-TDE~\cite{tang2020long} for causal inference.
EQL v2 achieves better results than all those methods. With ResNet-50-FPN backbone, it outperforms the winner of last year's challenge EQL by 4.3\%. We also compare the results under large models. With the same Cascade Mask R-CNN~\cite{cai2018cascade} framework equipped with ResNet-101-FPN, EQL v2 still has a 1.8\% higher AP than De-confound-TDE. With the same Hybrid Cascade R-CNN~\cite{chen2019hybrid} framework with ResNeXt-64x4d-FPN~\cite{xie2017resnext}, EQL v2 outperform BAGS by 0.8\% AP. Since LVIS v1.0 is recently proposed, not much work has reported their results on it. We mainly compare EQL v2 with De-confound-TDE. In addition, we also re-implement the equalization loss. The original EQL chooses a hyper-parameter $\lambda$ of $1.76 \times 10^{-3}$ for LVIS v0.5, we found this is not optimal for LVIS v1.0, so we tune this hyper-parameter and report the results with best value of $\lambda = 1.1 \times 10^{-3}$. The results are shown in Table \ref{tab:comparison_sota_lvis1}. Our method achieves higher overall AP across different backbones and frameworks. EQL v2 outperforms the De-confound-TDE by 6.3\% for rare categories.

\begin{figure*}[ht]
\begin{center}
\includegraphics[width=0.9\linewidth]{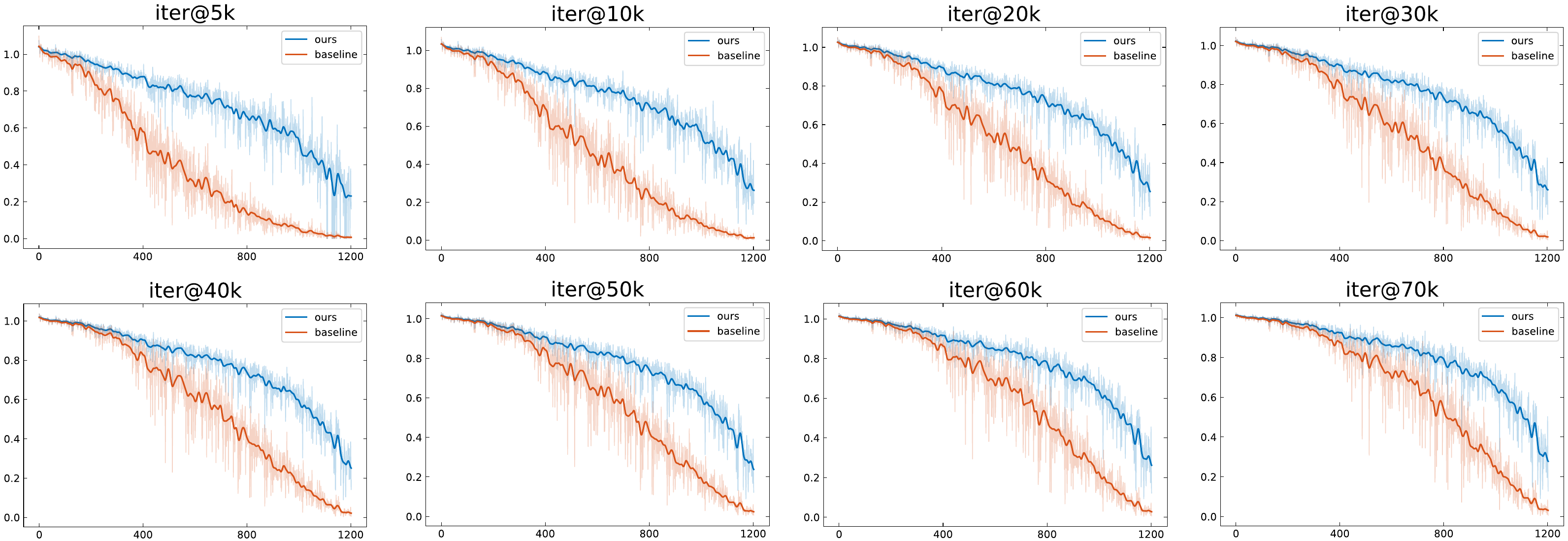}
\end{center}
   \caption{The accumulated gradients ratio of positives to negatives. Models are trained with total 75k iterations. We show the values at different training iteration. We compare the accumulated gradients of two models, Mask R-CNN with sigmoid loss and EQL v2.}
\label{fig:vis_balance_ratio}
\end{figure*}

\begin{figure}[t]
\begin{center}
\includegraphics[width=0.95\linewidth]{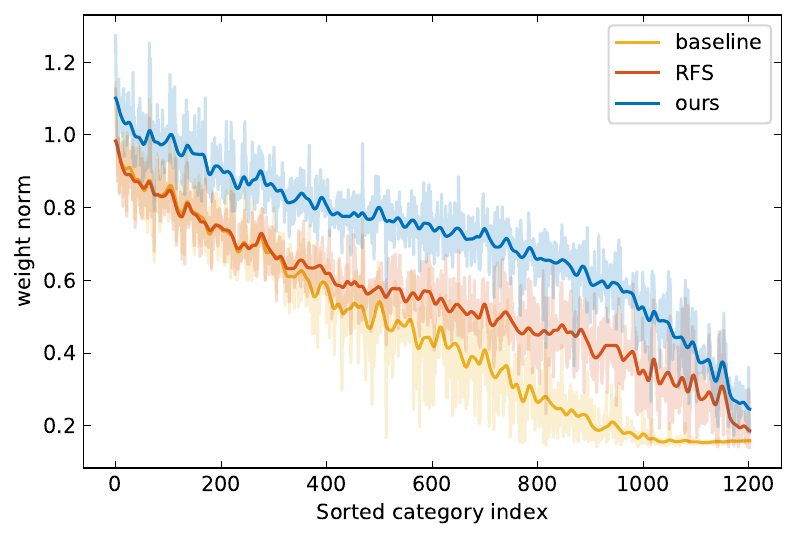}
\end{center}
   \vspace*{-1mm}
   \caption{The L2 weight norm of the fc-cls layer of models.}
\label{fig:vis_weight_norm}
\end{figure}

\subsection{Model Analysis}

\noindent \textbf{Do we have a more balanced gradient ratio?}  We visualize the gradient ratio of our method (Table \ref{tab:compare_e2e_decouple} (h)) and baseline model (Table \ref{tab:compare_e2e_decouple} (b)) during the training process, see Figure \ref{fig:vis_balance_ratio}. The baseline model does not have a balanced ratio for all categories. The positive gradients are overwhelmed by the negative gradients, especially for tail classes,  which makes it hard to detect them. And training longer does not help a lot. In contrast, EQL v2 preserves a more balanced gradient ratio in the entire training phase. 

\vspace{1mm} \noindent \textbf{Whether the classifiers are balanced?} Decoupled training methods \cite{kang2019decoupling, li2020overcoming} have shown that if models are trained with long-tailed distributed data, the weight norm in the last classifier layer~(fc-cls) is heavily biased. Those methods re-balance the classifier at the second fine-tuning stage, resulting in balanced weight norms. We also visualize the weight norm of the fc-cls of three models: baseline, RFS and EQL v2(Table \ref{tab:compare_e2e_decouple} (a), (c) and (h)), in Figure \ref{fig:vis_weight_norm}. The model trained with repeat factor sampling still suffers from biased weight norm. On the contrary, the model trained with EQL v2 has a more balanced weight norm.

\label{sec:better_repr}
\vspace{1mm} \noindent \textbf{Do we have a better representation?} To evaluate the quality of representations trained with our method. We adopt models trained with our method and standard training as pre-trained model. Then we follow the classic decoupled training recipe to re-train the classifier with frozen representations. The results are shown in Table \ref{tab:better_repr}. There are two main observations: Firstly, models initialized with EQL v2 always achieve a higher AP, resulting in 22.4 \vs 22.0 for cRT, 23.1 \vs 17.0 for LWS, 24.0 \vs 23.1 for BAGS. It verifies that we obtain a better representation by adopting EQL v2 compared to standard training. This result doubts the claim~\cite{kang2019decoupling} that re-weighing will hurt the representation. Secondly, the models get marginal improvements or even worse performance after decoupled training. The AP only increase 0.3 \% after using BAGS re-training, compared to the 23.7\% AP of EQL v2 (Table \ref{tab:compare_e2e_decouple}(h)), and AP drops 1.3 \% and 0.6 \% after using cRT and LWS respectively. It shows that decoupled training is not always necessary, we can train models with both a balanced classifier and better representations in an end-to-end manner.

\begin{figure}[t]
\begin{center}
\includegraphics[width=0.93\linewidth]{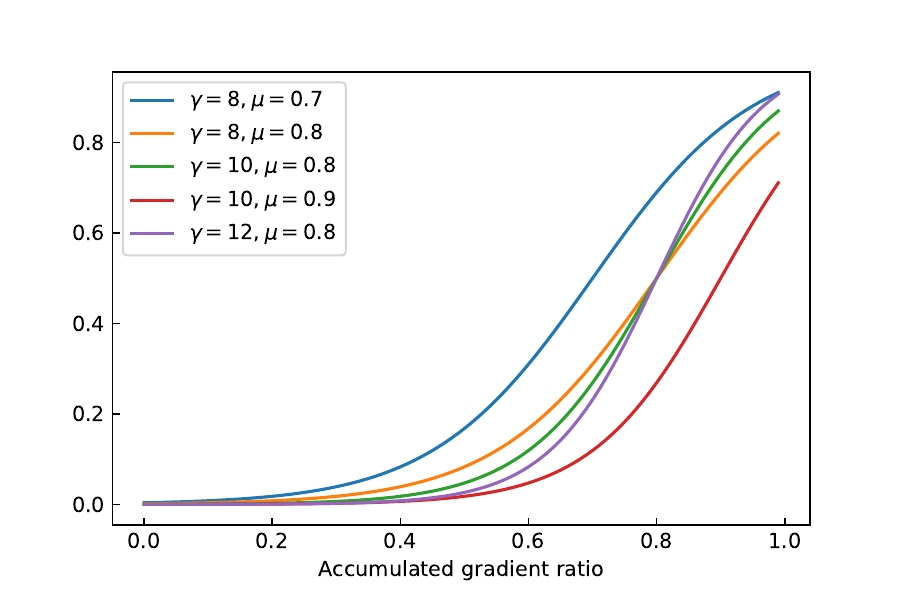}
\end{center}
   \vspace*{-2mm}
   \caption{Mapping functions with different $\mu$ and $\gamma$}
\label{fig:vis_mapping_func}
\end{figure}

\begin{table}
   \centering
   \setlength\tabcolsep{6pt}
   \begin{tabular}{c | c | c c c c c }
         method & EQL v2 & AP & AP\textsubscript{\textit{r}} & AP\textsubscript{\textit{c}} & AP\textsubscript{\textit{f}} & AP\textsubscript{\textit{b}}\\
         \thickhline
         cRT & & 22.0 & 13.5 & 20.8 & 27.1 & 22.1 \\
         cRT & \checkmark & 22.4 & 13.3 & 21.7 & 27.3 & 22.4 \\
         \hline
         LWS & & 17.0 & 2.0 & 13.5 & 27.4 & 17.5 \\
         LWS & \checkmark & 23.1 & 13.8 & 22.2 & 28.1 & 23.2 \\
         \hline
         BAGS & & 23.1 & 13.1 & 22.5 & 28.2 & 23.7 \\
         BAGS & \checkmark & 24.0 & 14.6 & 23.8 & 28.5 & 24.5 \\
   \end{tabular}
   \caption{Results of various decoupled training methods with different pre-trained models. EQL v2 indicates the pre-trained models are trained with EQL v2, otherwise with standard training. Only random samplers are used.}
   \label{tab:better_repr}
\end{table}

\begin{table}
   \centering
   \setlength\tabcolsep{7pt}
   \begin{tabular}{c c | c c c c c}
         $\gamma$ & $\mu$ &  AP & AP\textsubscript{\textit{r}} & AP\textsubscript{\textit{c}} & AP\textsubscript{\textit{f}} & AP\textsubscript{\textit{b}} \\
         \thickhline
         8 & 0.7 & 23.1 & 12.2 & 22.3 & 28.6 & 23.7 \\
         8 & 0.8 & 23.6 & 13.9 & 22.7 & \textbf{28.8} & 24.1 \\
         10 & 0.8 & 23.6 & 13.8 & \textbf{22.9} & 28.7 & \textbf{24.3} \\
         10 & 0.9 & 23.5 & 14.0 & 22.8 & 28.5 & 24.0 \\
         12 & 0.8 & \textbf{23.7} & 14.9 & 22.8 & 28.6 & 24.2  \\
         12 & 0.9 & 23.6 & \textbf{15.5} & 22.6 & 28.2 & 24.0 \\
   \end{tabular}
   \caption{Varying $\gamma$ and $\mu$ for mapping function. The positive up-weighting parameter $\alpha$ is set to 4. }
   \label{tab:hyper_mu_gamma}
\end{table}

\begin{table}
   \centering
   \setlength\tabcolsep{10pt}
   \begin{tabular}{c | c c c c c}
         $\alpha$ &  AP & AP\textsubscript{\textit{r}} & AP\textsubscript{\textit{c}} & AP\textsubscript{\textit{f}} & AP\textsubscript{\textit{b}} \\
         \thickhline
         0 & 19.7 & 7.3 & 17.6 & 27.6 & 20.5 \\
         1 & 21.9 & 11.3 & 20.7 & 28.0 & 22.8 \\
         2 & 23.0 & 13.5 & 22.0 & 28.4 & 23.7 \\
         4 & 23.7 & 14.9 & \textbf{22.8} & \textbf{28.6} & 24.2 \\
         8 & \textbf{24.0} & \textbf{16.5} & 22.7 & 28.6 & \textbf{24.4} \\
   \end{tabular}
   \caption{Varying $\alpha$. $\gamma$ and $\mu$ is set to 12 and 0.8 respectively.}
   \label{tab:hyper_alpha}
\end{table}

\subsection{Influence of hyper-parameters}
\label{sec:hyper_param} In Table~\ref{tab:hyper_mu_gamma} we investigate how the shape of mapping function $f(\cdot)$ in equation \ref{eq:map_func} affects the training. There are two hyper-parameters in the function, $\gamma$ and $\mu$. We can see that the detection result is not sensitive to the shape of the mapping function, and the AP increases stably when those two hyper-parameters move in a wide range. The visualization of mapping functions is shown in Figure~\ref{fig:vis_mapping_func}. The effect of $\alpha$ is present in Table~\ref{tab:hyper_alpha}. 

\subsection{C Sigmoid \textbf{\vs} 2C Softmax}
The idea of balancing gradient in EQL v2 is not restricted to sigmoid classifier. Recall that sigmoid uses a single output logit to represent each task and use the function $\sigma$ to estimate the probability. We show another choice of the classifier: 2C Softmax which uses 2 output logits for each task and adopts softmax function to estimate the probability. The extra output logit for each task can be regarded as a concept of \textit{others category} and introduces competition when doing inference so it helps reducing false-positives. In Table~\ref{tab:softmoid}, we compare the results of C-sigmoid and 2C-softmax under different settings. When only adding the objectiveness task and down-weighting the negative gradients, 2C-softmax achieves higher accuracy than C-sigmoid. These two designs reach comparable results after up-weighing the positive gradients.

\subsection{Experiments on Open Images Detection}
To verify the generalization ability to other datasets, we conduct experiments on the OpenImages~\cite{kuznetsova2018open}. OpenImages is another large-scale object detection dataset with long-tailed distributed categories. We use the data split of challenge 2019, which is a subset of OpenImages V5. The \texttt{train} set consists of 1.7M images of 500 categories. We evaluate our models on the 41k \texttt{val} set. In addition to the standard mAP@IOU=0.5 metric, we also group categories into five groups (100 categories per group) according to their instance numbers and report the mAP within each group respectively. The results are shown in Table~\ref{tab:openimage}. EQL v2 reaches an AP of 52.6, outperforming the baseline model and EQL by 9.5 AP and 7.3 AP respectively. For the tail group (AP1), the EQL v2 increases the AP by 22.3 point, which is much more than the improvement of EQL (6.4 AP). EQL v2 also outperform EQL considerably on the larger ResNet-101 backbone. For both baseline and EQL models, there is still a large performance gap between head and tail classes. EQL v2 brings all categories into a more equal status. It achieves similar accuracy for all categories groups. It is worth noting that we tune the hyper-parameter $\lambda$ in EQL which puts 250 categories into tail group for OpenImage. In contrast, the hyper-parameters of EQL v2 are kept the same as that on LVIS. Those experiments not only show the effectiveness but also good generalization ability of EQL v2. We also report the further tuned results of EQL v2 on Open Images. The values of $\mu$, $\gamma$ and $\alpha$ are 0.9, 12, 8 respectively.

\begin{table}
   \centering
    \small
    \setlength{\tabcolsep}{4pt}
   \begin{tabular}{l | c c c | c c c c}
         & obj? & neg? & pos? & AP & AP\textsubscript{\textit{r}} & AP\textsubscript{\textit{c}} & AP\textsubscript{\textit{f}} \\
         \thickhline
         C-sigmoid & \multirow{2}{*}{\checkmark} & & & 18.1 & 1.9 & 16.4 & 28.3\\
         2C-softmax & & & & \textbf{19.0} & \textbf{2.0} & \textbf{17.3} & \textbf{28.4} \\
         \hline
         C-sigmoid & \multirow{2}{*}{\checkmark} & \multirow{2}{*}{\checkmark} & & 19.7 & 7.3 & 17.6 & 27.6 \\
         2C-softmax & & & & \textbf{20.7} & \textbf{9.5} & \textbf{18.9} & \textbf{27.7} \\
         \hline
         C-sigmoid & \multirow{2}{*}{\checkmark} & \multirow{2}{*}{\checkmark} & \multirow{2}{*}{\checkmark} & 23.7 & \textbf{14.9} &\textbf{22.8} & 28.6 \\
         2C-softmax & & & & \textbf{23.7} & 14.9 & 22.7 & \textbf{28.7} \\
         
   \end{tabular}
   \caption{Comparison between C-sigmoid and 2C-softmax under different components.}
   \label{tab:softmoid}
\end{table}

\begin{table}
   \centering
   \small
   \setlength\tabcolsep{6pt}
   \begin{tabular}{l | c |  c c c c c}
          &  AP & AP1 & AP2 & AP3 & AP4 & AP5 \\ 
         \thickhline
         Faster-R50 & 43.1 & 26.3 & 42.5 & 45.2 & 48.2 & 52.6 \\
         EQL & 45.3 & 32.7 & 44.6 & 47.3 & 48.3 & 53.1 \\
         EQL v2\textsuperscript{\textdagger} & 52.6 & 48.6 & 52.0 & 53.0 & 53.4 & 55.8 \\
         EQL v2\textsuperscript{\textdaggerdbl} & \textbf{53.8} & \textbf{49.6} & \textbf{53.3} & \textbf{54.5} & \textbf{54.9} & \textbf{56.6} \\
         \hline
         Faster-R101 & 46.0 & 29.2 & 45.5 & 49.3 & 50.9 & 54.7 \\
         EQL & 48.0 & 36.1 & 47.2 & 50.5 & 51.0 & 55.0 \\
         EQL v2\textsuperscript{\textdagger} & 55.1 & 51.0 & 55.2 & 56.6 & 55.6 & 57.5 \\
         EQL v2\textsuperscript{\textdaggerdbl} & \textbf{55.6} & \textbf{51.5} & \textbf{55.5} & \textbf{57.5} & \textbf{55.8} & \textbf{57.6} \\
   \end{tabular}
   \caption{Results on \textbf{Open Images Challenge 2019} \texttt{val} set. The model Faster R-CNN~\cite{ren2015faster} with ResNet-FPN is trained with a schedule of 120k/160k/180k. Categories are grouped into five groups according to the instance number. AP1 is the AP of the first group, where categories have least annotations, AP5 is the AP of the last group, where categories have most annotations. \textdagger means that we directly use the hyper-parameters searched in LVIS, \textdaggerdbl means that we tune them in OpenImages.}
   \label{tab:openimage}
\end{table}

\section{Conclusion} 

In this work, we propose the key of improving performance for long-tailed object detection is to maintain balanced gradients between positives and negatives. An improved version of EQL, EQL v2 is then proposed to dynamically balance the gradient ratio between positives to negatives in the training phase. It brings large improvements with notably boosting on tail categories across various frameworks. As an end-to-end training method, it beats all existing methods on the challenging LVIS benchmark, including the dominant decoupled training schema. 

\clearpage
\begin{appendices}

\section{Mapping Function Types} 
In Table~\ref{tab:mapping_func}, we make the comparison among several variant mapping functions. Results show that our proposed sigmoid-like mapping function achieves the highest AP.

\begin{table}[h]
  \small
  \centering
  \setlength\tabcolsep{2.5pt}
  \begin{tabular}{l | c c c c | c c c c }
         \hline
         & \multicolumn{4}{|c|}{neg \checkmark} & \multicolumn{4}{c}{neg \checkmark pos \checkmark} \\
         \hline
         type & AP & AP\textsubscript{\textit{r}} & AP\textsubscript{\textit{c}} & AP\textsubscript{\textit{f}} & AP & AP\textsubscript{\textit{r}} & AP\textsubscript{\textit{c}} & AP\textsubscript{\textit{f}} \\
         \hline
         sqrt ($y=\sqrt{x}$)& 18.4 & 2.1 & 17.6 & 28.2 & 21.4 & 6.2 & 21.0 & 28.6 \\
         linear ($y = x$)& 18.8 & 2.0 & 16.9 & 28.3  & 22.6 & 10.0 & 22.0 & 28.7 \\
         exp ($y = x^2$)& 19.1 & 2.1 & 17.6 & 28.2  & 23.2 & 11.9 & 22.7 & 28.8 \\
         \hline
         ours & 19.7 & 7.3 & 17.6 & 27.6 & 23.7 & 14.9 & 22.8 & 28.6  \\
  \end{tabular}
  \caption{Comparison between different mapping functions}
  \label{tab:mapping_func}
\end{table}

\section{More Ablations of Hyper-Params} 
We have conducted more ablation studies of $\mu$ and $\gamma$, and the experiment results are presented in Table~\ref{tab:hyper_mu_gamma}. Since $\mu$ represents the value which we think as a high enough gradient ratio, lowering its values significant degrades the accuracy. It is better to choose a higher value for it, \eg, 0.8, 0.9. $\gamma$ is more robust when $\mu$ is in a reasonable range.

\begin{table}[ht!]
  \small
  \centering
  \setlength\tabcolsep{8pt}
  \begin{tabular}{c c| c c c c }
         $\gamma$ & $\mu$ &  AP & AP\textsubscript{\textit{r}} & AP\textsubscript{\textit{c}} & AP\textsubscript{\textit{f}} \\
         \hline
         10 & 0.4 & 20.4 & 5.7 & 19.0 & 28.4  \\
         10 & 0.5 & 21.6 & 9.6 & 20.1 & 28.5  \\ 
         10 & 0.6 & 22.4 & 11.8 & 21.1 & 28.6 \\
         5 & 0.8 & 23.1 & 12.4 & 22.4 & 28.7  \\
         15 & 0.8 & 23.2 & 15.0 & 22.0 & 28.1 \\
         20 & 0.8 & 22.3 & 15.4 & 20.6 & 27.3 \\
  \end{tabular}
  \caption{More Ablations of $\gamma$ and $\mu$. }
  \label{tab:hyper_mu_gamma}
\end{table}

\end{appendices}

{\small
\bibliographystyle{ieee_fullname}
\bibliography{egbib}
}

\end{document}